\documentclass[journal]{IEEEtran}
\usepackage[latin1]{inputenc}
\usepackage[dvips]{graphicx}

\usepackage{algorithm}
\usepackage{algorithmic}
\usepackage{color}
 \usepackage{amsmath,amsfonts}
\begin{document}

\title{Conceptualization of seeded region growing  by pixels aggregation. Part 1: the framework }

\author{Vincent Tariel }

\maketitle
\begin{abstract}
Adams and Bishop have proposed in 1994 a novel region growing algorithm called seeded region growing by pixels aggregation (SRGPA). This paper introduces a framework to implement an algorithm using SRGPA. This framework is built around two concepts: localization and organization of applied action. This conceptualization gives a quick implementation of algorithms, a direct translation between the mathematical idea and the numerical implementation, and an improvement of algorithms efficiency. 
\end{abstract}

\begin{keywords}
 Minkowski addition, seeded region growing by pixel aggregation. 
\end{keywords}
\IEEEpeerreviewmaketitle

\section{Introduction}
Many fields in computer science, stereovision \cite{KANADE1994}, mathematical morphology \cite{Serra1982},
use algorithm which principle is Seeded Region Growing by Pixels Aggregation (SRGPA). This method consists in initializing each region with a seed, then processing pixels aggregation on regions, iterating this aggregation until getting a nilpotence \cite{ADAMS1994}\cite{Hojjatoleslami1998}. The general purpose of this field is to define a metric divided into two distinct categories \cite{Ballard1982}: the region feature like the tint \cite{ADAMS1994} and region boundary discontinuity\cite{Beucher1979}. The localization, the actualization and the organization of action in the interest zone is done manually.\\ 
In this paper, we propose a conceptualization of localization and organization of action in the interest zone. First, a new mathematical object, Zone of Influence (ZI), is associated to each region to define a zone on the region outer boundary. Second as long as the algorithm is processed, the regions (de)grow, implying some ZI fluctuations. A procedure is defined to actualize efficiently the ZI such that some operations have to be computed only on the (de)growth neighbourhood. Third, in order to organize action on pixels belonging to the ZI, a system of queues (SQ) is defined to sort out each pixel belonging to the ZI depending on a metric and the entering time. Fourth, a library named Population has been created to implement these three first mathematical parts. An algorithm implementation using SRGPA with this library is quick and efficient. The translation between the mathematical idea and the numerical implementation is direct. Last, since this library has been optimized mathematically and numerically, each algorithm created with its is also optimized. In the article number
\section{Conceptualization of interest zone}
The basic idea is to define three objects: Zone of Influence (ZI), System of Queues (SQ) and Population. The algorithm implementation using SRGPA is focused on the utilisation of these three objects. An object ZI is associated to each region and localizes a zone on the outer boundary of its region. For example, a ZI can be the outer boundary region excluding  all other regions. An algorithm using SRGPA is not global (no treatment for a block of pixels) but local (the iteration is applied pixel by pixel belonging to the ZI). To manage the pixel by pixel organisation, a SQ sorts out all pixels belonging to ZI depending on the metric and the entering time. It gives the possibility to select a pixel following a value of the metric and a condition of the entering time. The object population links all regions/ZI and permits the (de)growth of regions. A pseudo-library, named Population, implements these three objects. An algorithm can be implemented easier and faster with this library, fitted for SRGPA. The utility of this library will be developped in the article number 3 of this serie.\\ 
This section is decomposed in four subsections: localization, actualization, organization, implementation.
\subsection{Localization}
\begin{figure}[h]
    \begin{center}
\includegraphics[width=1.6cm]{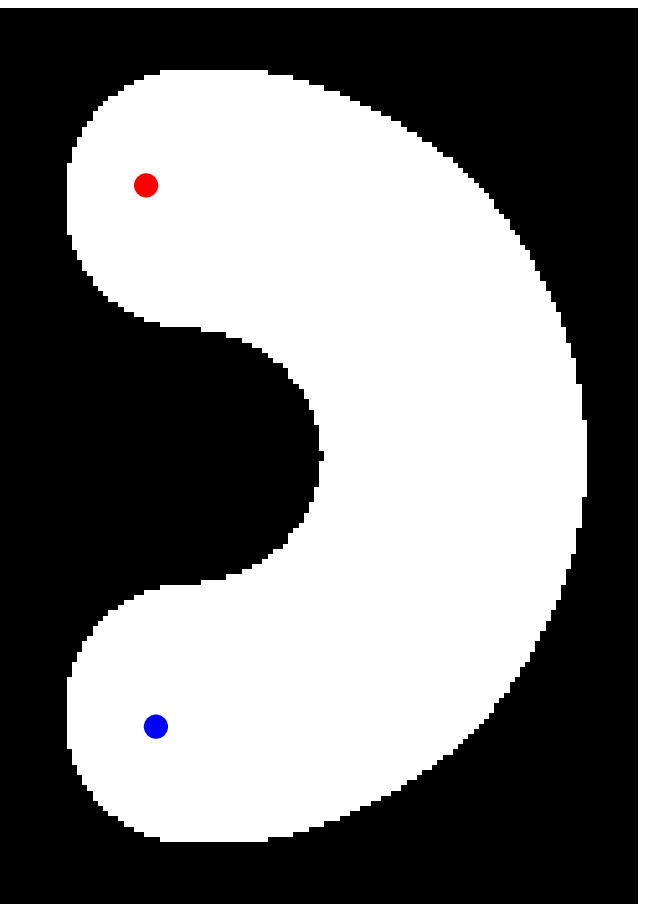}
\includegraphics[width=1.6cm]{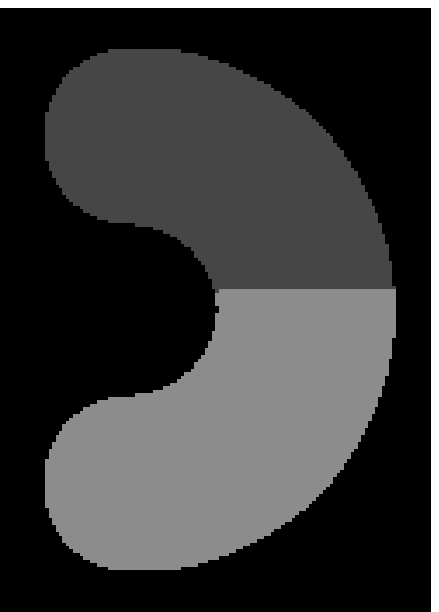}
\caption{Left figure: the white pixels are the desert island, the black pixels are the water and the red and blue pixels are the seeds of the two lichens. Right figure: after an infinite time, the lowest grey-level represents the blue lichen region and the middle grey-level represents the red lichen region.} 
        \label{beam}
\end{center}
\end{figure}

The aim of the object Zone of Influence (ZI) is to localize the zone of interest where the algorithm is processing operations. Its mathematical definition respects three conditions: simple, generic and numerically translatable.\\
To understand the definition of this object, an example is given. Let assumed that it exists two types of lichens with two different colours, blue and red. At time $t=0$, there is only a seed of each lichen on a desert island surrounded by water (see figure~\ref{beam}). At each step of time, each lichen grows on each pixel $x$ belonging to its boundary if $x$ doesn't belong to the other lichen or water. The region water never grows. The object ZI defines the localization of the growth for each region.\\
Let $X_{red}^t$, $X_{blue}^t$, $X_{water}^t$ three subsets of the space $E$ representing the red and blue lichens and the water. We define the object, ZI, for each region, $Z_{red}^t$, $Z_{blue}^t$, $Z_{water}^t$.\\
First the boundary can be defined as:
\begin{eqnarray*}
Z^t_{red}=&(X_{red}^t \oplus V)\setminus (X^t_{red})\\
Z^t_{blue}=&(X_{blue}^t \oplus V)\setminus (X^t_{blue})\\
Z^t_{water}=&(X_{red}^t \oplus \emptyset)= \emptyset \\
&\mbox{The water region never grows}
\end{eqnarray*}
The symbol $\oplus$ means the Minkowski addition: $X\oplus V = \cup_{x\in X}V(x)$. $V$ is an application of $E$ to $\mathcal{P}(E)$, for example a ball.\\ 
We add to this first definition the property that each pixel of the object ZI doesn't belong to the water and the other lichen.
\begin{eqnarray*}
Z^t_{red}=&(X_{red}^t \oplus V)\setminus (X^t_{red}\cup X^t_{water} \cup X^t_{blue})\\
Z^t_{blue}=&(X_{blue}^t \oplus V)\setminus (X^t_{blue}\cup X^t_{water} \cup X^t_{red})\\
Z^t_{water}=&(X_{red}^t \oplus \emptyset)= \emptyset \\
&\mbox{The water region never grows}
\end{eqnarray*}  
This last definition defines correctly the initial constraints.\\
 In a general case, let $\mathcal{X}^t=\{X_0^t,\ldots,X_n^t\}$ a set of regions. So for each region, we define a neighbourhood function, $V_i=V \mbox{ or }\emptyset $\footnote{$\emptyset$ means an application of $E$ to $\emptyset$} and a restricted set $N_i\subset \mathbb{N}_n$ such that each ZI associated to a region is defined as:
\begin{eqnarray*}
Z_i^t &= & (X_i^t \oplus V_i)\setminus(\bigcup\limits_{j \in N_i} X^t_{j})
\end{eqnarray*} 
In the next papers\cite{Tariel2008d}\cite{Tariel2008f}, we will define differently the $V_i$ and $N_i$ to localize correctly the different zones of interest for each algorithm. This object ZI is generic and its mathematical definition is simple. The last property, numerically translatable, is presented in the next subsection.
\subsection{Actualization}
\label{toto}
Firstly, we introduce two operators in the set theory, addition and subtraction.\\
The addition, $+$, between two sets $A$ and $B$, is the disjoint union: $A+B=A\uplus B$\\
The subtraction\footnote{The addition is commutative, not the subtraction}, $-$, between two sets $A$ and $B$, is: $A-B=(A^c+B)^c$\\
We have defined these operators because they are useful in the numerical conversion of mathematical expressions.\\
The SRGPA implies some fluctuations (growth or degrowth) of the regions. As the ZI depend on regions, the ZI have to be actualized. For the actualization, the simplest way is to calculate the fluctuation for each ZI. At each region pixel (de)growth, the numerical cost is high: the complexity is $\Theta(n)$ where n is the ZI/regions number.  Numerically, the actualization has to be fast.\\
A solution is to reformulate the link between ZI and the regions: the ZI at time t+1 depends on ZI at time t plus or minus a set defined in the neighbourhood of the fluctuation. This reformulation gives an actualization with a constant complexity\footnote{This section is quite technical and it is not necessary for this article understanding}.\\
In this subsection, we suppose that we have one and only one region fluctuation $A^t$\footnote{Except the initialisation with seeds, each region fluctuation is a single pixel $A^t=\{x\}$} between two steps of time:
\begin{eqnarray*}
(\forall t \in \mathbb{N}, \exists ! i \in \mathbb{N}_n)\begin{cases}X_i^{t+1}=X_i^{t}\pm A^t \\ \forall j \in \mathbb{N}_n\setminus i, X_j^{t+1}=X_j^{t})\end{cases}
\end{eqnarray*} 
We see  two parts in the expression of ZI: $Z_i^t = \overbrace{ (X_{i,m}^t \oplus V_i)}^{\mbox{myself}}\setminus(\overbrace{\bigcup\limits_{j \in N_i} X^t_{j,o}}^{\mbox{other}})$. The "myself part" depends only on the associated region. The "other part" depends on all regions. This is the reason why we decompose the actualization in two steps. First, we actualize the "myself part": $X^{t+1/2}_{i,m}=X^{t}_{i,m}\pm A^t$ then the "others part": $X^{t+1}_{i,o}=X^{t+1/2}_{i,o}\pm A^t$ (see figure~\ref{actu}). All the proofs are in the appendice \ref{demos}.\\
\subsubsection{myself part}
We suppose that there is only the fluctuation of $X^{t}_{i,m}$ between time $t$ and $t+1/2$.\\ 
For all j different of i, $Z_j^{t+1/2}$ is equal to $Z_j^{t}$ because they do not depend on $X^{t}_{i,m}$.\\
If it is a growth $X_{i,m}^{t+1/2}=X_{i,m}^t+A^t$ then
\begin{eqnarray*}
Z_i^{t+1/2}&=&Z^t_{i}+ (A^t\oplus V_i)\setminus Z^t_{i}\setminus (\bigcup\limits_{j \in N_i} X^{t}_{j,o})
\end{eqnarray*}
If it is a degrowth $X_{i,m}^{t+1/2}=X_{i,m}^t-A^t$ then
\begin{eqnarray*}
Z_i^{t+1/2}&=&Z_i^t - ((A^t\oplus V_i)\setminus(X_{i,m}^{t+1/2}\oplus V_i))\setminus (Z_i^t)^c
\end{eqnarray*}
\subsubsection{other part} 
We suppose that there is only the fluctuation of $X^{t}_{i,o}$ between time $t+1/2$ and $t+1$\\ 
If it is a growth $X_{i,o}^{t+1}=X_{i,o}^{t+1/2}+A^t$ then
\begin{eqnarray*}
\forall j \in \mathbb{N}_n\begin{cases} Z_j^{t+1}=Z^{t+1/2}_j - A^t\setminus (Z^{t+1/2}_j)^c 
\\ \hspace{2cm} \mbox{ if } (i \in N_j)\wedge (V_j\neq \emptyset) \\ Z_j^{t+1}=Z^{t+1/2}_j \mbox{ else}\end{cases}
\end{eqnarray*}
The symbol $\wedge$ means the \textbf{and} in the symbolic logic.\\
If it is a degrowth $X_{i,o}^{t+1}=X_{i,o}^{t+1/2}-A^t$ then
\begin{eqnarray*}
\forall j \in \mathbb{N}_n\begin{cases} Z_j^{t+1}=Z_j^{t+1/2}+\\(((A^t\setminus (X^{t+1/2}_{m,j}\oplus V)^c)\setminus  (\bigcup\limits_{k \in N_j} X^{t+1}_{k,o}))\setminus Z_j^{t+1/2})\\ \hspace{2cm}\mbox{ if } (i \in N_j) \wedge (V_j\neq \emptyset) \\ Z_j^{t+1}=Z_j^{t+1/2}  \mbox{ else}\end{cases}
\end{eqnarray*}
\begin{figure} 
\begin{center}
\includegraphics[width=1.5cm]{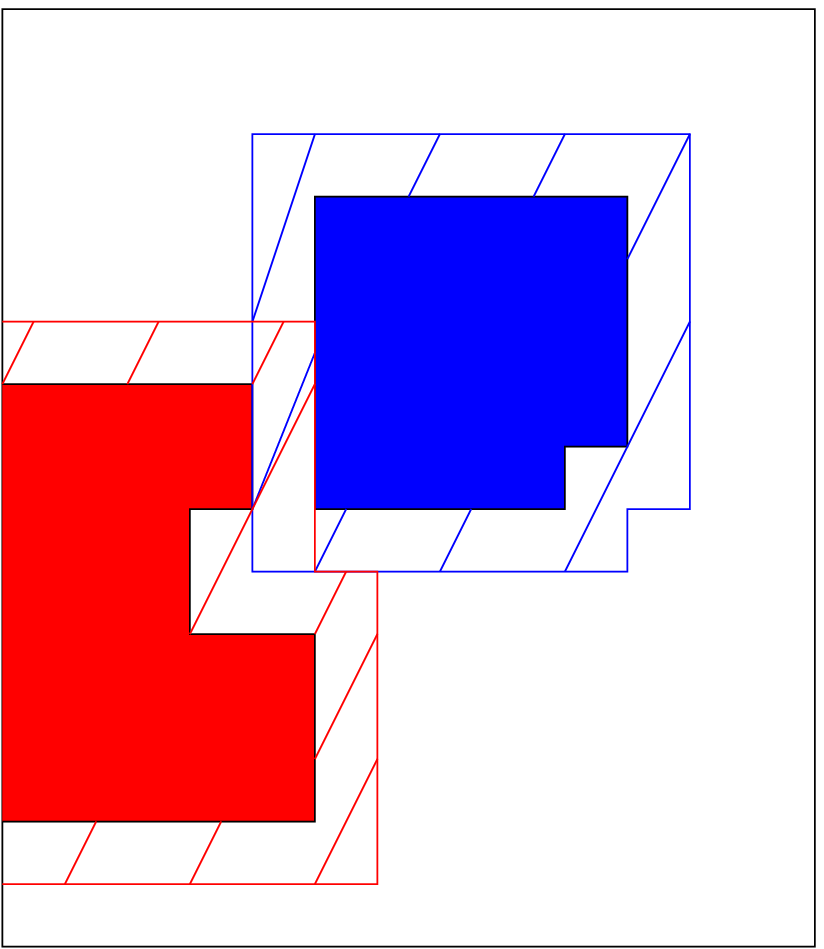}
\includegraphics[width=1.5cm]{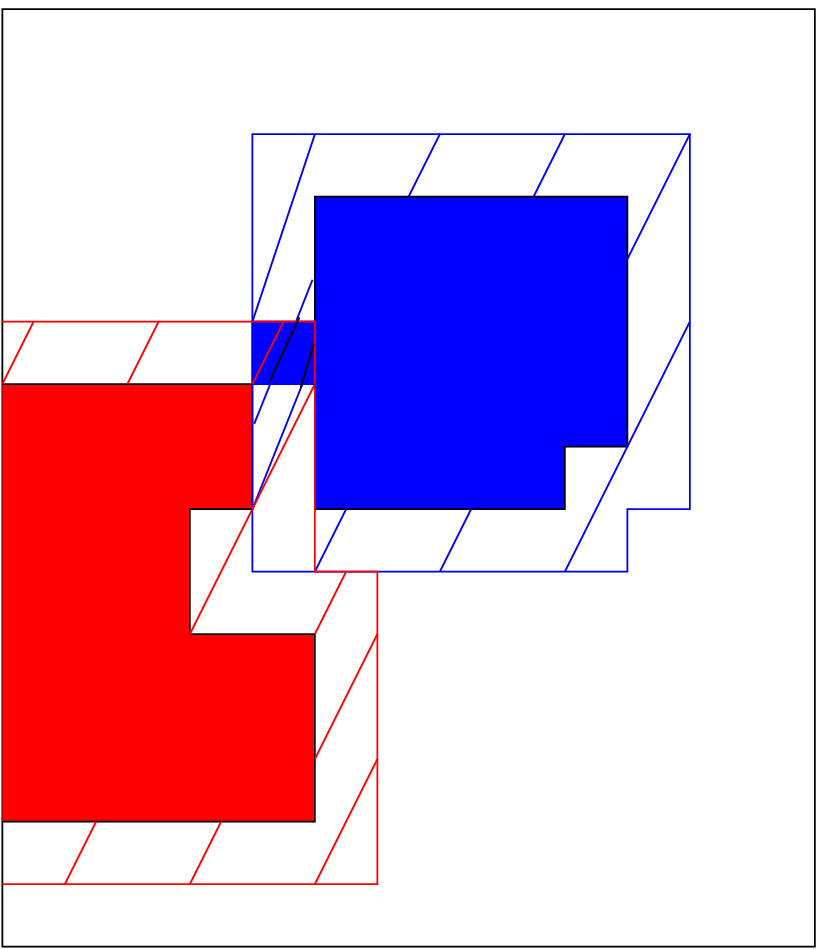}
\includegraphics[width=1.5cm]{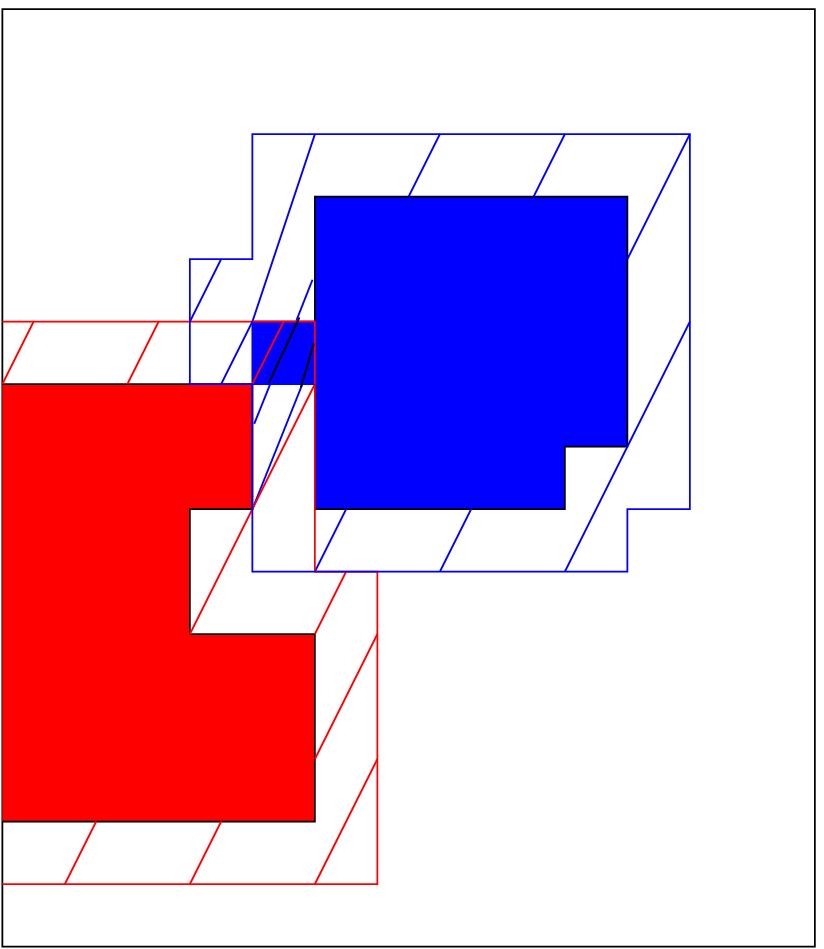}
\includegraphics[width=1.5cm]{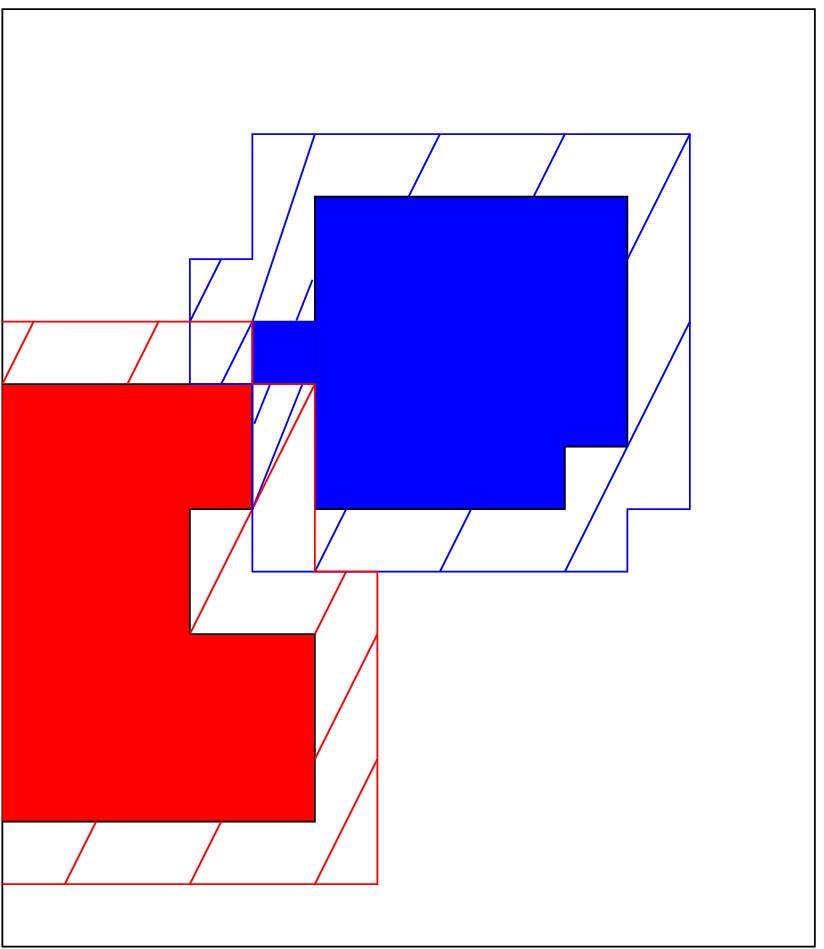}\\
\includegraphics[width=1.5cm]{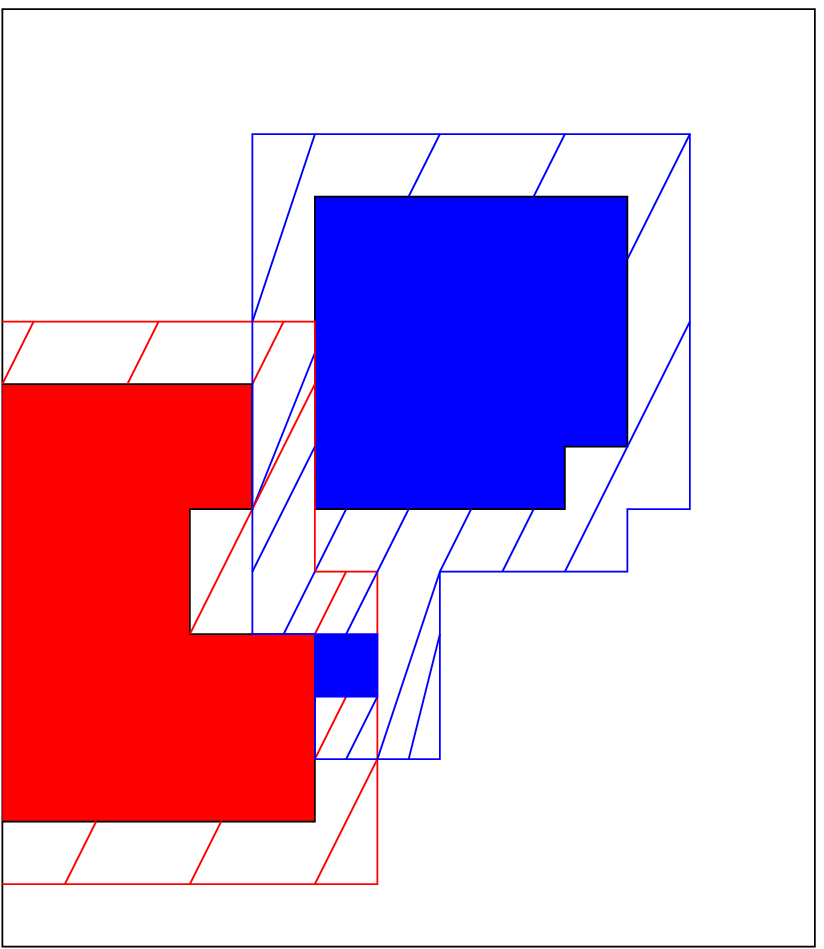}
\includegraphics[width=1.5cm]{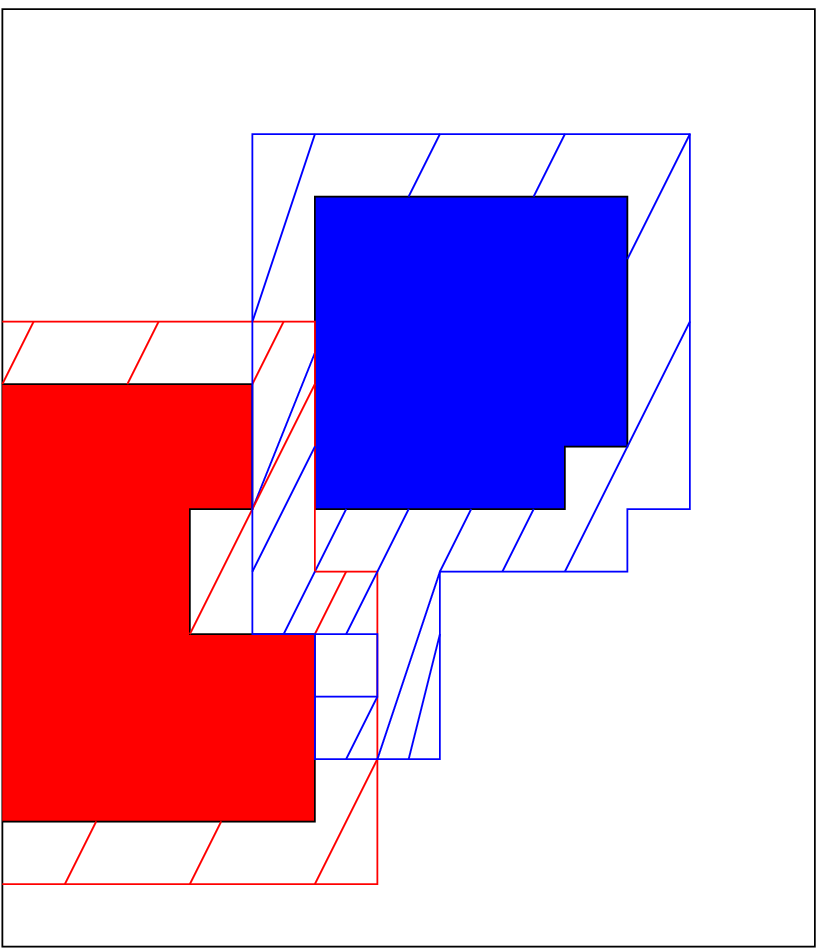}
\includegraphics[width=1.5cm]{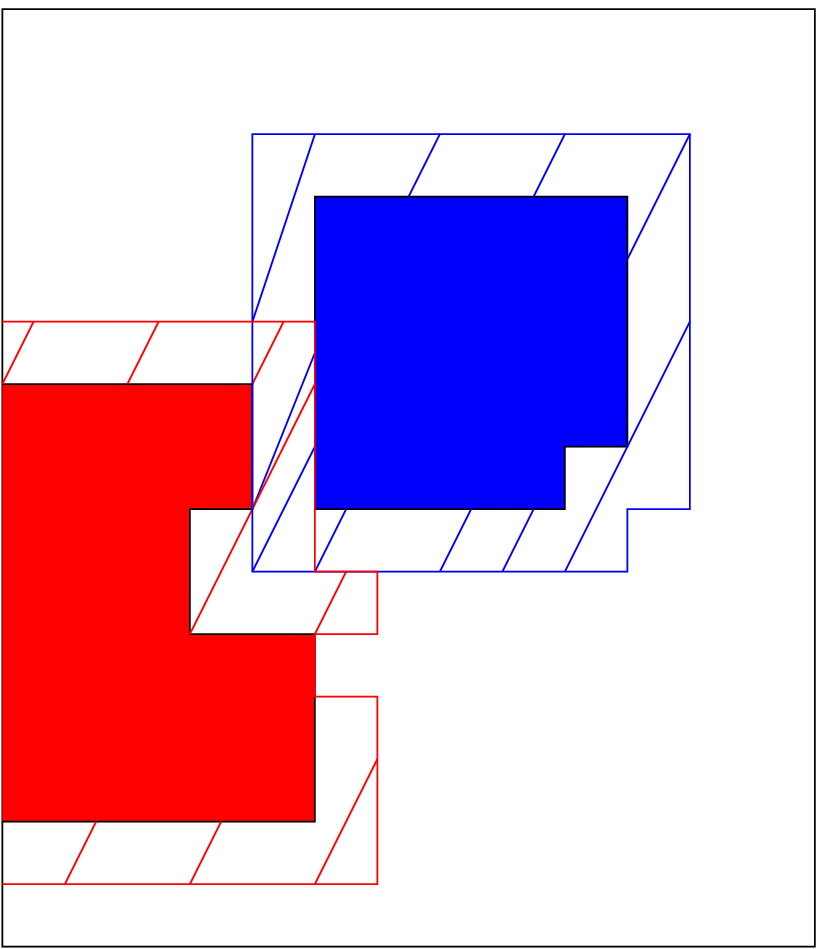}
\includegraphics[width=1.5cm]{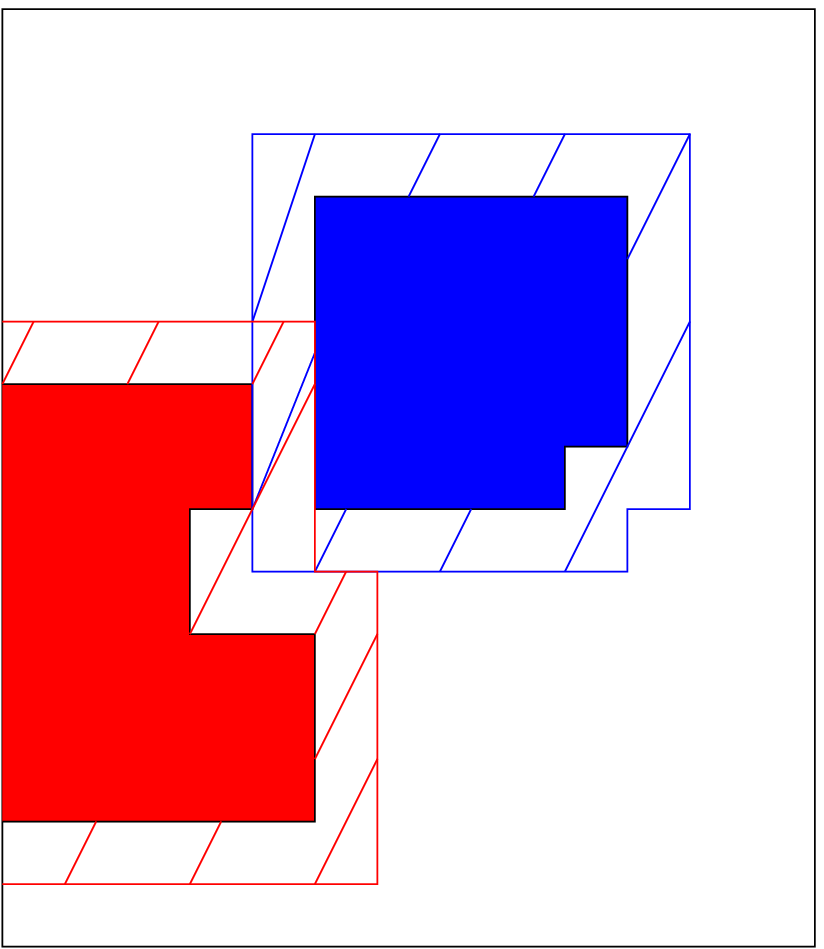}
\caption{Two regions: $X_r^t$ and $X_b^t$, two ZI $Z_r^t=(X_r^t \oplus V_r)\setminus (X^t_r \cup X^t_b)$ and $Z_b^t=(X_b^t \oplus V_b)\setminus (X^t_r \cup X^t_b)$. Both series show the ZI actualization after a growth for the first serie and degrowth for the second serie. The first figure is the initial state, the second figure is the (de)growth of the blue region, the third figure is the actualization of the "myself part" of the ZI, the last figure is the actualization of the "other part" of the ZI.}
\label{actu}
\end{center}
\end{figure}
\subsubsection{Numerically}
If we have a growth $A^t$ of the region $X^t_i$, the ZI are actualized in two steps.\\
First, the "myself part" actualization: $Z_i^{t+1/2}=Z^t_{i}+ (A^t\oplus V_i)\setminus Z^t_{i}\setminus (\bigcup\limits_{j \in N_i} X^{t}_{j,o})$
\begin{algorithmic}[20]
\FORALL{$\forall x \in A^t\oplus V_i$}
\IF{$x\notin Z^t_{i}$}
\IF{$x\notin \bigcup\limits_{j \in N_i} X^{t}_{j,o}$}
\STATE $Z_i^{t+1/2}=Z_i^{t}+x$;
\ENDIF
\ENDIF
\ENDFOR
\end{algorithmic}
Second, the "other part" actualization:
\begin{eqnarray*}
\forall j \begin{cases} Z_j^{t+1}=Z^{t+1/2}_j - A^t\setminus (Z^{t+1/2}_j)^c \mbox{ if } (i \in N_j)\wedge (V_j\neq \emptyset) \\ Z_j^{t+1}=Z^{t+1/2}_j \mbox{ else}\end{cases}
\end{eqnarray*}
\begin{algorithmic}[20]
\FORALL{$\forall x \in A^t$}
\FORALL{$\forall j :  x\in Z_j^{t+1/2}$}
\IF{$i\in N_j\wedge V_j\neq \emptyset $}
\STATE $Z_j^{t+1}=Z_j^{t+1/2}-x$;
\ENDIF
\ENDFOR
\ENDFOR
\end{algorithmic}
If we have a degrowth $A^t$ of the region $X^t_i$, the ZI are actualized in two steps.\\
First, the "myself part" actualization: $Z_i^{t+1/2}=Z_i^t - ((A^t\oplus V_i)\setminus(X_{i,m}^{t+1}\oplus V_i))\setminus (Z_i^t)^c$
\begin{algorithmic}[20]
\FORALL{$\forall x \in A^t\oplus V_i$}
\IF{$x\in Z_i^t$}
\IF{$ x\oplus V_i^{-1} \cap X_{i,m}^{t+1} = \emptyset $}
\STATE $Z_i^{t+1/2}=Z_i^{t}-x$;
\ENDIF
\ENDIF
\ENDFOR
\end{algorithmic}

Second, the "other part" actualization:\footnote{ The assumption that for all j, $V_j$ is equal to $\emptyset $ or $V$ , is crucial here.}
\begin{eqnarray*}
\forall j \begin{cases}  Z_j^{t+1}=Z_j^{t+1/2}+\\(((A^t\setminus (X^{t+1/2}_{m,j}\oplus V)^c)\setminus  (\bigcup\limits_{k \in N_j} X^{t+1}_{k,o}))\setminus Z_j^{t+1/2}) \\\hspace{2cm} \mbox{ if } (i \in N_j) \wedge (V_j\neq \emptyset) \\ Z_j^{t+1}=Z_j^{t+1/2}  \mbox{ else}\end{cases}
\end{eqnarray*}
\begin{algorithmic}[20]
\FORALL{$\forall x \in A^t$}
\FORALL{$\forall j :  (x\oplus V^{-1}) \cap X^{t+1/2}_{m,j} \neq \emptyset $}
\IF{$i\notin \bigcup\limits_{j \in N_i} X^{t+1}_{j,o}$}
\IF{$i\notin Z_j^{t+1/2} $}
\IF{$i\in N_j\wedge V_j\neq \emptyset $}
\STATE $Z_j^{t+1}=Z_j^{t+1/2}+x$;
\ENDIF
\ENDIF
\ENDIF
\ENDFOR
\ENDFOR
\end{algorithmic}
The complexity of this actualization is time constant $\Theta(1)$ if two data structures are implemented:
\begin{eqnarray*}
L_X(x)=\{\forall i : x \in X_i\}\\
L_Z(x)=\{\forall i : x \in Z_i\}
\end{eqnarray*}
At this step, the ZI localize correctly the zone of interest and there is a procedure to actualize the ZI efficiently. The next subsection will organize the action.

\subsection{Organization}
The general purpose of seed region growing is to define a metric\cite{Ballard1982} applied to the pixels belonging to the ZI. The priority order of pixel by pixel process depends on this metric and the entering time. At each iteration of pixel by pixel, a pixel is chosen because it respects some constraints on the metric and the entering time. The System of Queue (SQ) manages this organisation.\\
The set $B^t$ is the couples to add to the SQ between the $t+1$ and $t$: 
\[B^t=\{ (x,i) :  x \notin Z_{i}^t \wedge  x \in Z_{i}^{t+1}\}\]
The set $C^t$ is the couples to substract to the SQ between the $t+1$ and $t$: 
\[C^t=\{ (x,i) :  x \in Z_{i}^t \wedge  x \notin Z_{i}^{t+1}\}\]   
The SQ is like a "factory store". There are:
\begin{itemize}
\item the products, $B^t$, to store and to give a label,
\item the products, $C^t$, to remove from the store, 
\item a customer that chooses product  depending on its label and its entering time. 
\end{itemize}
To get the best efficiency in the "factory store" organisation, the three following points must be respected:
\subsubsection{Data structure}
A data structure is a way to store data in a computer. A data structure depends on the operations to be performed, using as few resources both execution time and memory space, as possible. Generally\cite{Tariel2008d}, the data structure used is composed by $n$ queues, $Q_n^t=\{q_0^t,\ldots,q_{n}^t\}$, because the strategy is:
\begin{itemize}
\item to store each couple $(x,i)$ whose metric is the same in the queue (a queue is associated to a quantification of the metric and $n$ is the metric range), 
\item to organize in First In First Out (FIFO) "what comes in first is handled first, what comes in next waits until the first is finished".
\end{itemize}
But the organisation can be a random access to handle randomly a couple\cite{Tariel2008f}. 
\subsubsection{The ordering attribute function}
\begin{figure}
\begin{center}
\includegraphics[width=3cm]{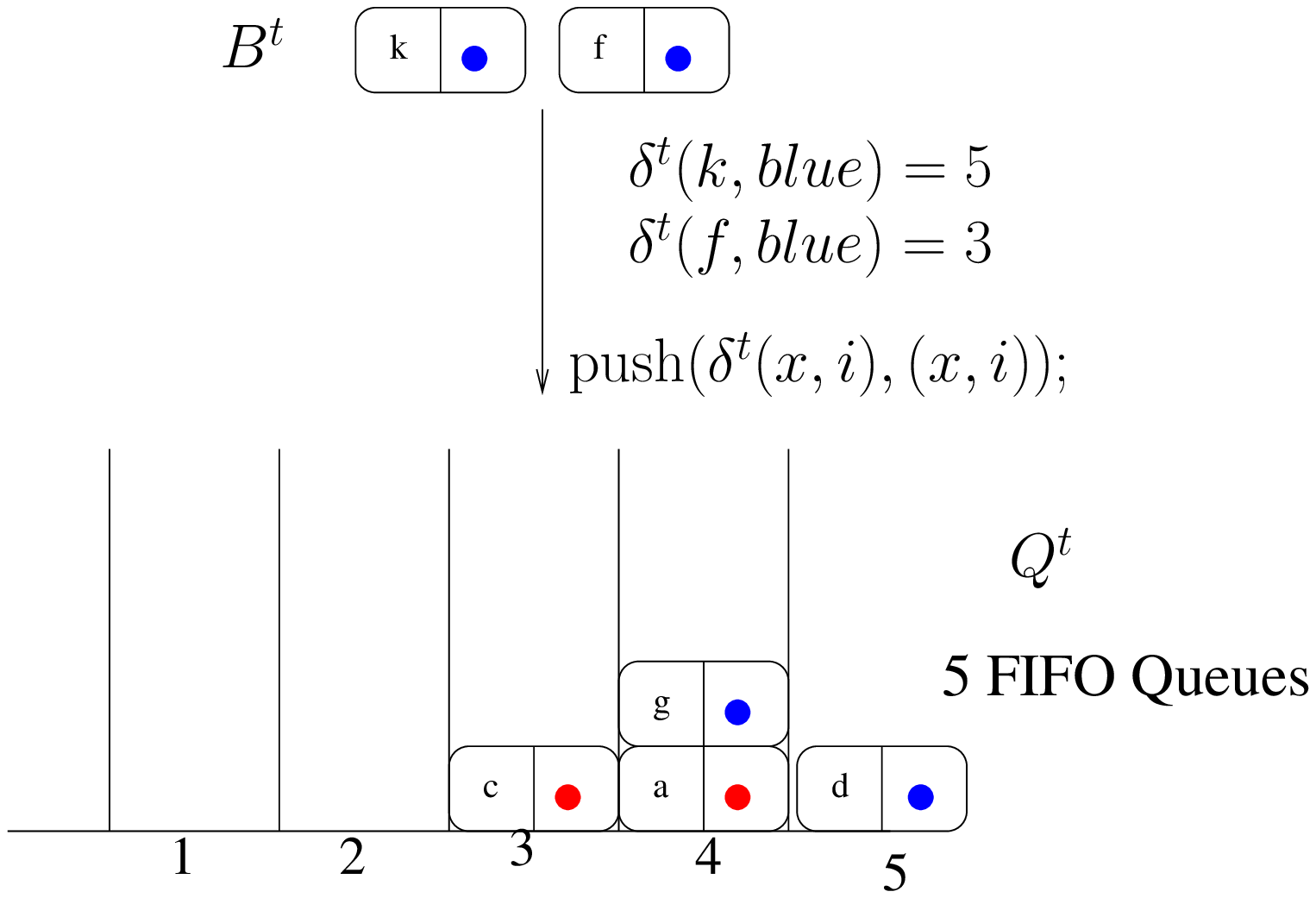}
\includegraphics[width=3cm]{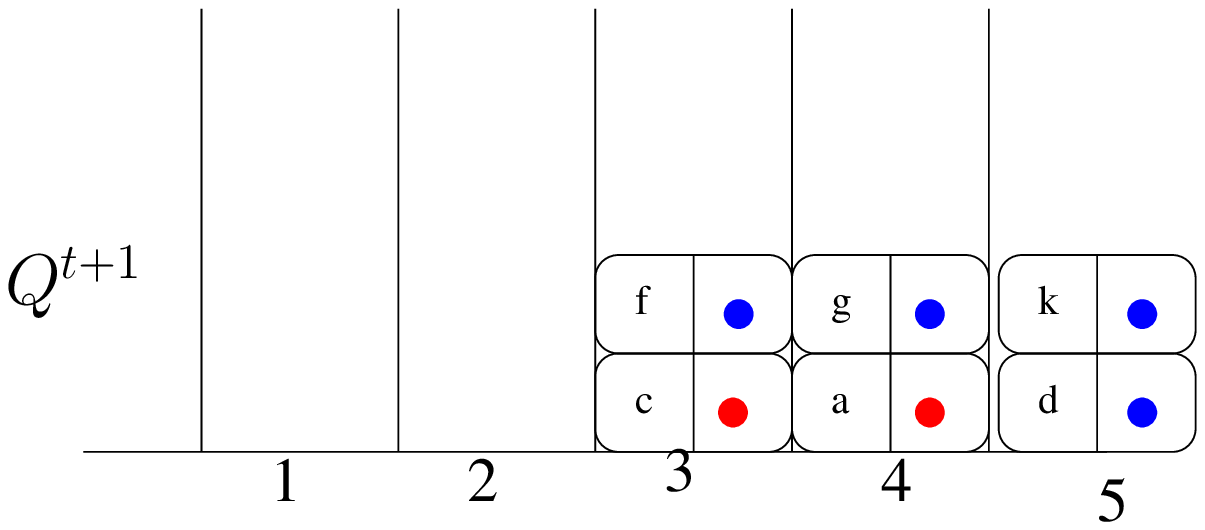}
\caption{Evolution of $Q^t$ after the addition of the set $B^t$  }
\label{inf3:fig}
\end{center} 
\end{figure}

The aim of the ordering attribute function, $\delta$, is to assign  each couple $(x,i)$ of $B^t$ in the appropriate queue. The label $k$ of the queue is given by $k=\delta(x,i)$ (see figure~\ref{inf3:fig}). NB: if the label $k$ is equal to OUT then the couple is not affected to the set of queues.
\subsubsection{The extraction function}
\begin{figure}
\begin{center}
\includegraphics[width=3cm]{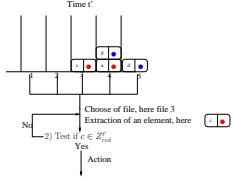}
\caption{How to act? 1)Choose a queue 2)Extract a couple $(x,i)$ 3)Test if $x\in Z_i^{t'}$ 4) Yes, action, 5) No, come back to the first step}
\label{inf4:fig}
\end{center} 
\end{figure}

Between time $t$ and $t+1$, all couple belonging to $C^t$ have to be removed  in the set of queue. The numerical cost of this strategy is important. Another solution is to do nothing except when an couple $(x,i)$ is extracted to a queue $j$ at time $t$, then the test $x\in Z_i^{t}$ is operated\footnote{When a couple $(x,i)$ enters in the SQ, $x$ belongs to $Z^{t-t'}_i$ but is it still the case, now?}. If the test returns true, operations are processed on this couple. If the test returns false, then this couple is deleted. So the extraction function has two components: choose of file and test  (figure \ref{inf4:fig}). 
The general definition of the two objects, ZI and SQ, is finished and the next subsection introduces the algorithm implementation.

\subsection{The algorithm implementation}
In this section, a pseudo-library, called Population, is presented\footnote{Free software available at http://pmc.polytechnique.edu/$\sim$vta/population.zip}.
This pseudo-library is composed with three objects. Each object is used in the elaboration of an algorithm. These three objects are:
\begin{itemize}
\item the SQ that manages the "store factory".
\item the tribe that  collects a restricted set and a neighbourhood to define after a ZI associated to a region.
\item the population that collects the set of regions/ZI and gives method for regions (de)growing.
\end{itemize}
 The  implementation of this library is done using the C++ language. The conceptualization and the generic programming\footnote{An algorithm is not attached to a type of data but to some data properties.} permit:
\begin{enumerate}
\item  an implementation of algorithms using SRGPA with less than fourty lines of codes,
\item the application of these algorithms whatever the dimension of the image (principally 2D, 3D) and the type of pixel/voxel,
\item the optimization of all algorithms using SRGPA. Since the library has been optimized, all algorithms using this library will benefit from the optimization.  
\end{enumerate}
\textbf{Object: SQ}\\
The aim of the SQ is to manage the "store factory".
Its implementation implies the definition of:
\begin{enumerate}
\item the ordering attribute function, $\delta (x, i)$, to give the label of the queue where a couple will be stored, for example:
\begin{itemize}
\item $\delta (x, i)=0$
\item $\delta (x, i)=\mathcal{I}(x)$ with $\mathcal{I}$  an image (an application of $E$ to $\mathbb{N}_n$)
\item $\ldots$
\end{itemize}
\item the data structure of the elementary queue, (queue, linked list,$\ldots$)
\item the number of queues.
\end{enumerate}
This class provides three methods:
\begin{enumerate}
\item select$\_$queue(int i): selects the queue $i$ (the customer selects a product with the label $i$).
\item pop(): returns couple $(x,i)$ from the selected queue (the customer gets a product with the label $i$).
\item empty(): returns true if the selected queue is not empty, false else (the customer asks if there is still product with the label $i$). 
% \item push(int j, couple $(x,i)$), push the couple $(x,i)$ to the queue j (the customer gives a product to the store factory and numerically.
\end{enumerate}
\textbf{Object: Tribe}\\
The aim of the object tribe is to collect a restricted set\footnote{Numerically, the object restricted set, $N$, has two fields:
\begin{itemize}
\item  $a\_w$ (all or without),  a boolean
\item  $L$,  a list of integer 
\end{itemize}
if
\begin{eqnarray*}
N=& \begin{cases} \mathbb{N}_n\setminus L \mbox{ if } a\_w=true\\  L \mbox{ else } \end{cases} 
\end{eqnarray*}}, $N$, and a neighbourhood, $V$ to define a ZI associated to a region. This object doesn't provide method but is used by the object Population.\\ 
\textbf{Object: Population}\\
The object Population allows the creation of the set regions/ZI: $(\mathcal{X}^t,\mathcal{Z}^t)_n=\{(X^t_0,N_0,V_0,Z_0^t),\ldots,(X^t_n,N_n,V_n,Z_n^t)\}$ such that $Z_i^t=(X_i^t \oplus V_i)\setminus(\bigcup\limits_{j \in N_i} X^t_{j})$. The initial state of the object population is $(\mathcal{X}^{t=0},\mathcal{Z}^{t=0})_{n}=\emptyset$. So its methods permit the addition of region/ZI and the addition/subtraction of  pixel(s) to a region. 
\begin{enumerate}
\item growth$\_$tribe( Tribe tr=$(V, N)$):
\begin{itemize}
\item creates an empty region/ZI, $(X,Z)^t$ with $Z^t=(X^t \oplus V)\setminus(\bigcup\limits_{j \in N} X^t_{j})$,
\item pushes this region/ZI at the end of the set of regions/ZI, $(\mathcal{X}^{t+1},\mathcal{Z}^{t+1})_{n+1}=\{ (\mathcal{X}^t,\mathcal{Z}^t)_{n}, \{ (X^t_{n+1}=\emptyset,V, N, Z^t_{n+1})\}\}$
\item returns the label of the created region/ZI in the set of regions/ZI.
\end{itemize}
\item growth(Pixel x, int label$\_$region/ZI): adds the pixel x to the region i, actualizes all the ZI and stores the set $B^t$ in the SQ using the ordering attribute function.
\item growth(Set A, int label$\_$region/ZI): adds the set A to the region i, actualizes all the ZI and stores the set $B^t$ in the SQ using the ordering attribute function.
\item degrowth(Pixel x, int label$\_$region/ZI): subtracts the pixel x for all regions $i$ if x belongs to $X_i$, actualizes all the ZI and stores the set $B^t$ in the SQ using the ordering attribute function.
\item Z()[x] gives the list $Z_x=\{\forall i : x\in Z_i\}$
\item X()[x] gives the list $X_x=\{\forall i : x\in X_i\}$
\end{enumerate}

\begin{figure}
\begin{center}
\includegraphics[width=1.6cm]{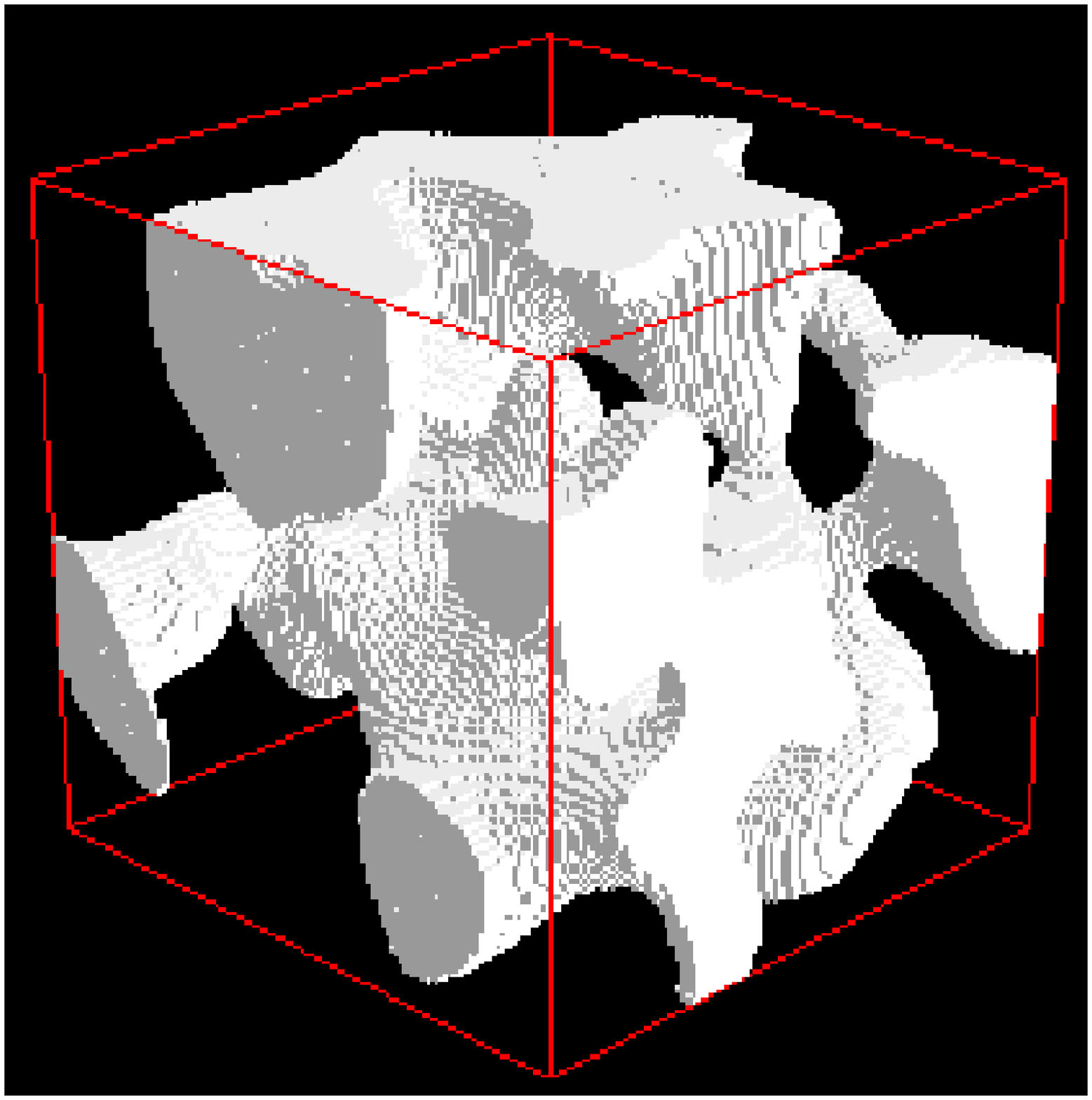}
\includegraphics[width=1.6cm]{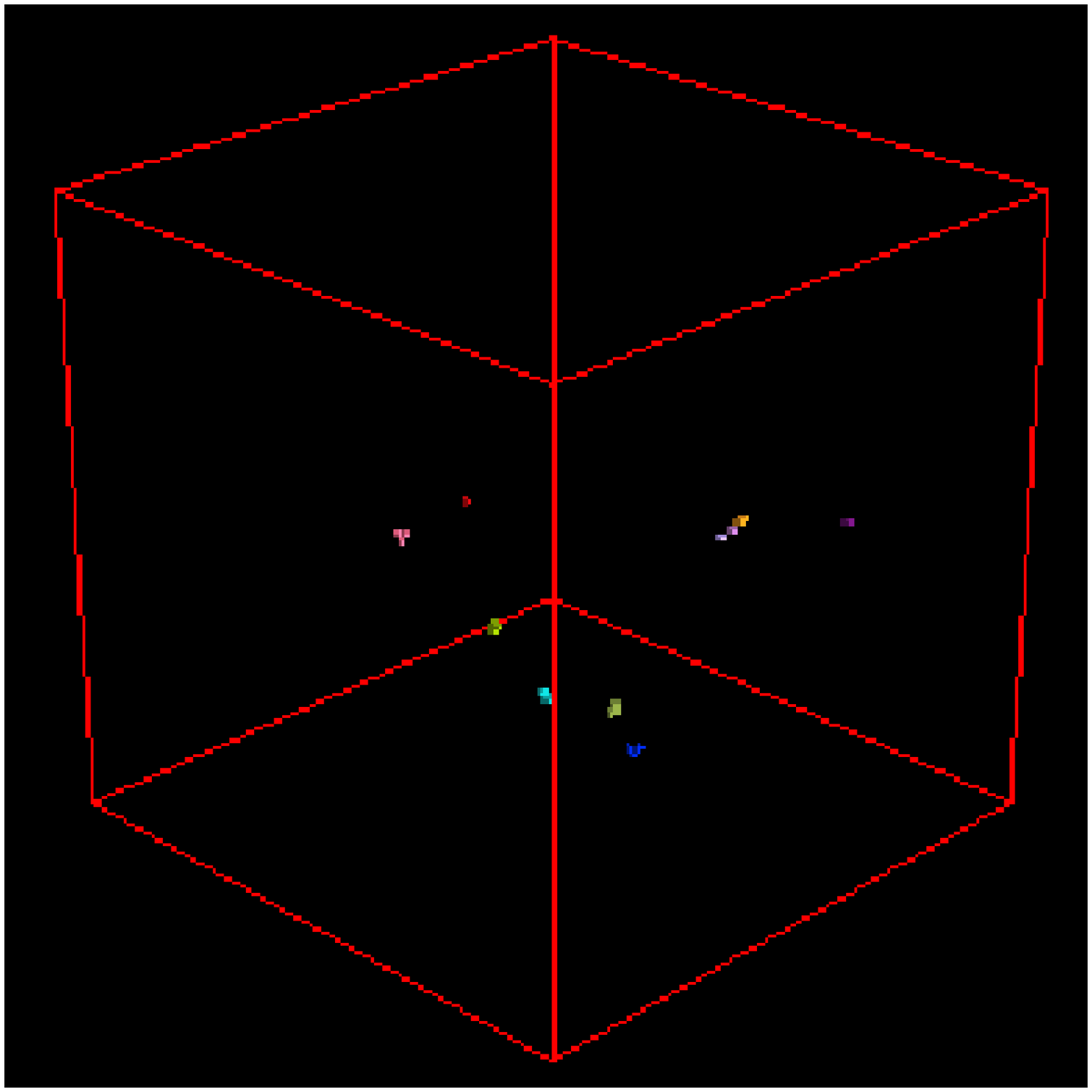}
\includegraphics[width=1.6cm]{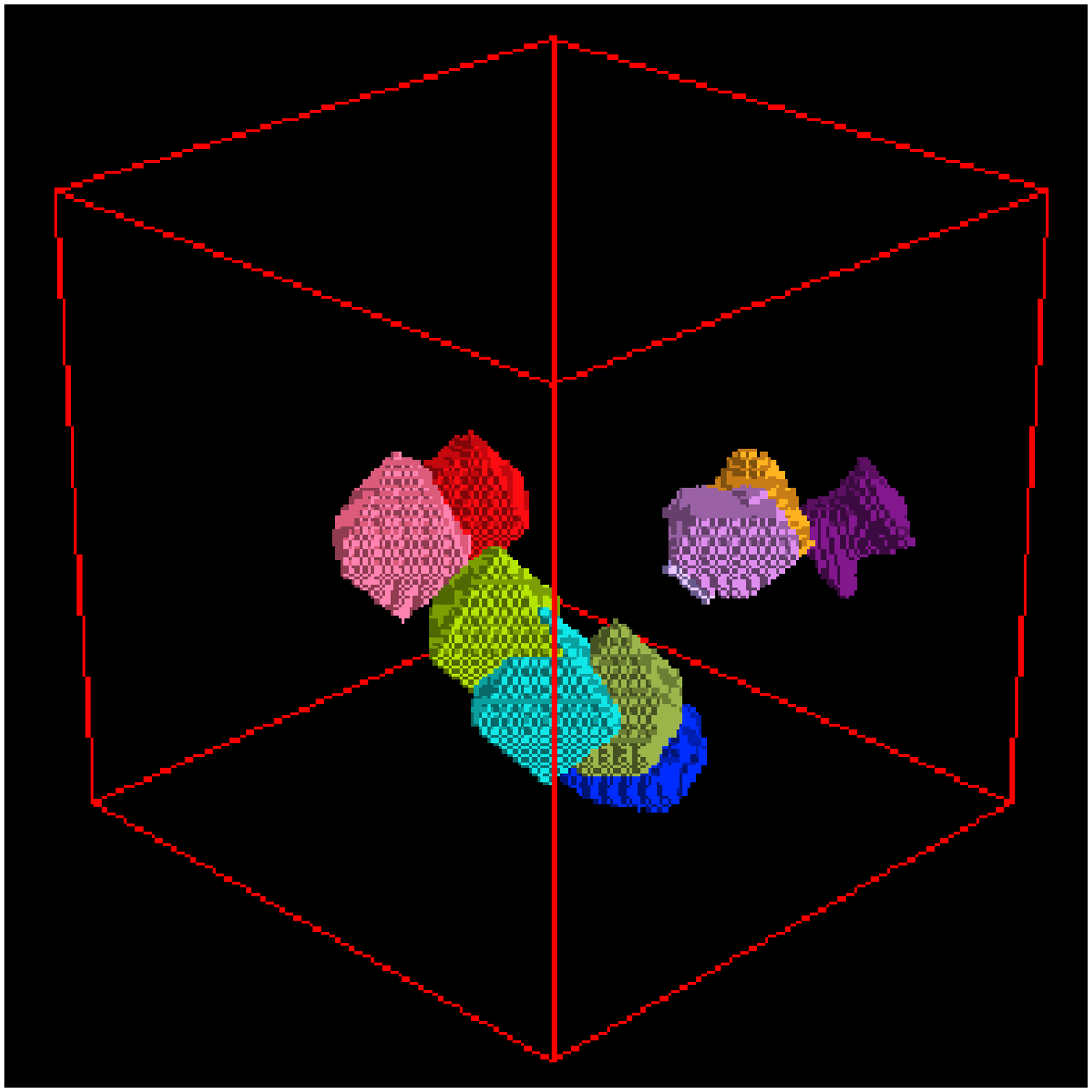}
\includegraphics[width=1.6cm]{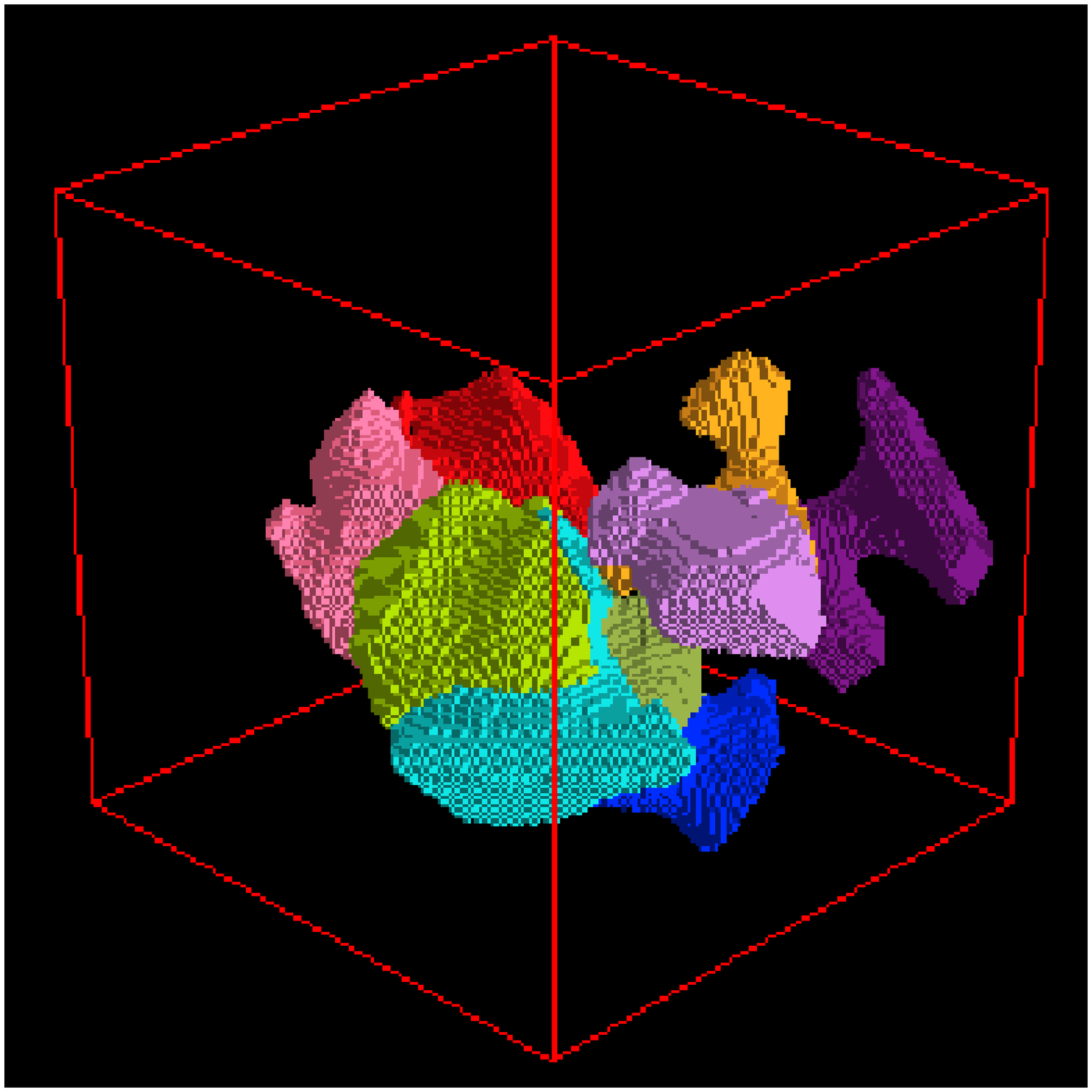}
\includegraphics[width=1.6cm]{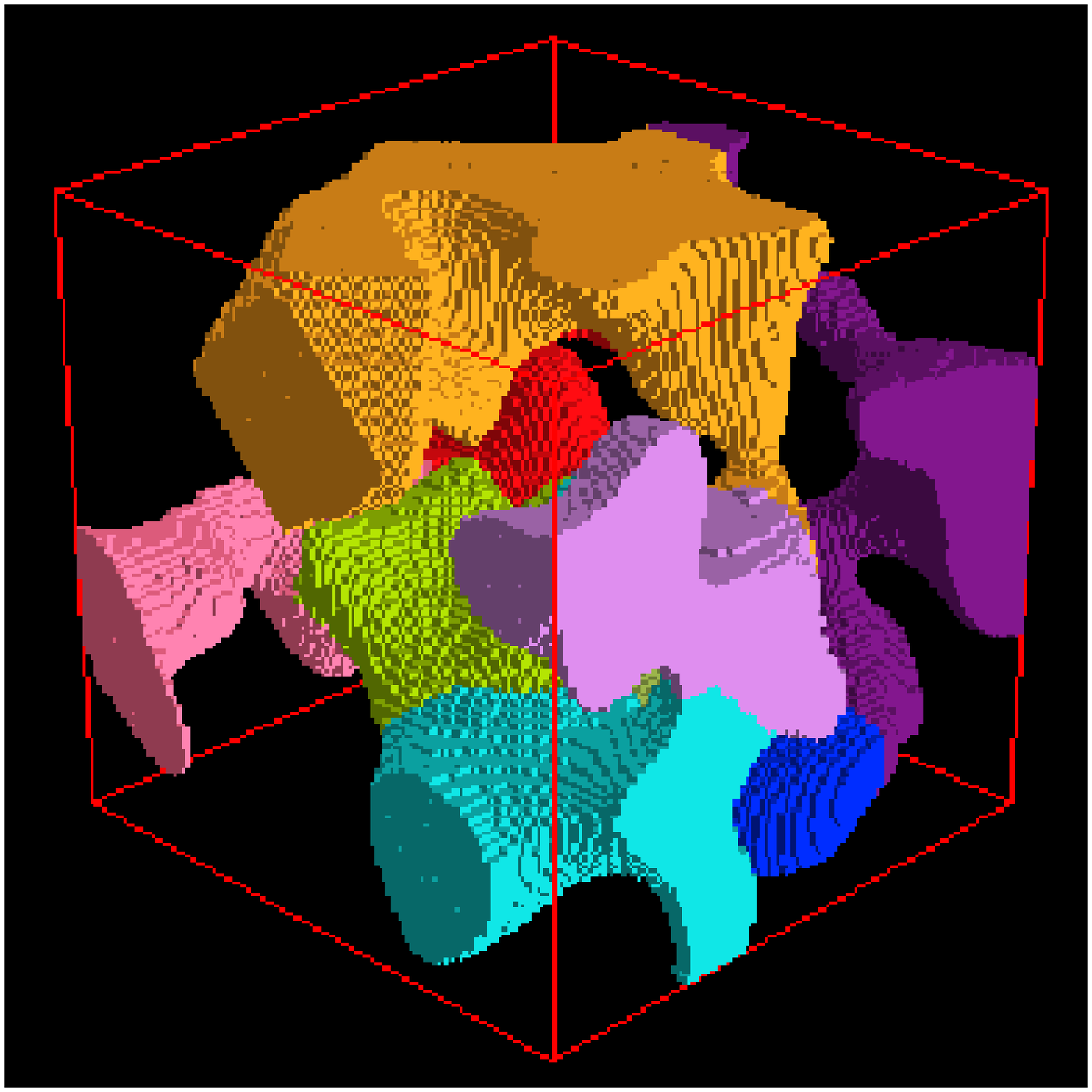}
\caption{The first figure is the desertic island, the second figure is the set of seeds, the next figures are the geodesic dilatation such as each colour represent a specefic region. A video is  available at http://pmc.polytechnique.edu/$\sim$vta/geodesic.mpeg}
\label{vicord}
\end{center} 
\end{figure}

\textbf{Algorithmic in general}\\
An algorithm is usually presented with:
\begin{enumerate}
\item Definition of SQ: the ordering attribute function $\delta$, the elementary queue $q$ and the number of queues
\item Definition of the different tribes
\item Creation of regions/ZI using these tribes
\item Initialization of the regions/ZI with seeds
\item The growing process:
 \begin{algorithmic}[20]
\STATE Select a queue
\WHILE{the selected queue is not empty}
\STATE Extract a couple $(x,i)$ from the selected queue
\STATE Process some operations like (de)growth$\_$pixel(x,i) 
\ENDWHILE
\end{algorithmic}
\item Return result
\end{enumerate}
The algorithm~\ref{alg0} (see figure~\ref{beam} and ~\ref{vicord}) is an example of these steps linking.
 \begin{algorithm}[h!tp]
\caption{the growth of lichens in a desertic island (geodesic dilatation)}
\label{alg0}
\algsetup{indent=1em}
\begin{algorithmic}[20]
 \REQUIRE $I$, $S$ , $V$ \textit{//The binary image ($I(x)\neq 0$ means $x$ belongs to the desertic island and else belongs to the water), the seeds, the neighbourhood}
\STATE // \textbf{initialization}
\STATE System$\_$Queue s$\_$q( $\delta(x,i)=0\mbox{ if }I(x)\neq 0, OUT\mbox{ else}$, FIFO, 1); \textit{//A single FIFO queue such as the pixel belonging to the water are not pushed in the SQ.}
\STATE Population p (s$\_$q); \textit{//create the object Population}
\STATE Restreint $N=\mathbb{N}$; 
\STATE Tribe actif(V, N);
\FORALL{$\forall s_i\in S$} 
\STATE int ref$\_$tr   = p.growth$\_$tribe(actif); \textit{//create a region/ZI, $(X^t_i,Z^t_i)$ such as $Z^t_i=(X_{i}^t \oplus V)\setminus (\bigcup\limits_{j \in \mathbb{N}} X_{j})$}
\STATE  p.growth($s_i$, ref$\_$tr ); 
\ENDFOR
\STATE // \textbf{the growing process}
\STATE  s$\_$q.select$\_$queue(0); \textit{//Select the single FIFO queue.}
\WHILE{s$\_$q.empty()==false}
\STATE  $(x,i)=s$\_$q$.pop();
\STATE p.growth(x, i ); 
\ENDWHILE
\RETURN p.X();
\end{algorithmic}
 \end{algorithm}
\section{Conclusion}
In this paper, we have conceptualized the localization and the organization of seed region growing method by pixels aggregation.\\
In the conceptualization part, we define two objects and one procedure to make possible the creation of the library, called Population. The first object, zone of influence, is associated to each region to localize a zone on the outer boundary region. The second object, the system of queue, organizes the queue of pixel by pixel aggregation around a concept of "factory store". The procedure is defined to actualize at time constant the zones of influence after a fluctuation of a region. This procedure and these objects permit to this library to be numerically efficient and to implement algorithm faster.\\
This first paper will be follow by others. In the paper part 2, we will present a procedure to define a boundary between the region such as its localization
does not depend on the seeded region initialisation order. In the paper part 3, this library will be used to implement a wide range of algorithms. In the paper part 4, a simple, generic and robust method will be proposed to extract the components from experimental tridimensionnal  images of granular materials and porous media obtained by X-ray tomography using  algorithms coming from SRGPA. In the paper part 5, the cutting of the porous media in elementary pore will be done following two different conventions: morphological and topological. In the paper part 6, we will propose an efficient procedure to reconstruct a three-dimensional (3D) medium using morphological information: chord length distribution and two-point correlation. The efficient is due to the localization of the interchange procedure in the boundary of the phases.

\appendices

\section{Proof of the acualization}
\label{demos}
\subsection{Growth of the myself region}
If there is only the growth of the myself region: $X_{i,m}^{t+1/2}=X_{i,m}^t+A^t$ then
\begin{eqnarray}
Z_i^{t+1/2}&=&Z^t_{i}+ (A^t\oplus V_i)\setminus Z^t_{i}\setminus (\bigcup\limits_{j \in N_i} X^{t}_{j,o})
\end{eqnarray}
\textbf{Proof:}\\
\begin{eqnarray*}
Z_i^{t+1/2}&= & (X_{i,m}^{t+1/2} \oplus V_i)\setminus (\bigcup\limits_{j \in N_i} X^{t+1/2}_{j,o})
\end{eqnarray*}
As we have only the growth of the myself region: $X_{i,m}^{t+1/2}=X_{i,m}^t+A^t$ then
\begin{eqnarray*}
Z_i^{t+1/2}&= & ((X_{i,m}^t+A^t) \oplus V_i)\setminus (\bigcup\limits_{j \in N_i} X^{t}_{j,o})
\end{eqnarray*}
As $(A\cup B)\oplus V = (A\oplus V)\cup(B\oplus V)$ so with $R_i^t=\bigcup\limits_{j \in N_i} X^{t}_{j,o}$
\begin{eqnarray*}
Z_i^{t+1/2}&=&((X_{i,m}^{t}\oplus V_i)\cup(A^t \oplus V_i))\setminus R_i^t
\end{eqnarray*}
We know that $A\cup B = A + (B\setminus A)$:
\begin{eqnarray*}
Z_i^{t+1/2}&=&((X_{i,m}^{t}\oplus V_i)+(A^t\oplus V_i\setminus (X_{i,m}^{t}\oplus V_i)))\setminus R_i^t
\end{eqnarray*}
and $(A+B)\setminus C= (A\setminus C)+(B\setminus C)$, then
\begin{eqnarray*}
Z_i^{t+1/2}&=&(X_{i,m}^{t}\oplus V_i)\setminus R_i^t+(A^t\oplus V_i\setminus (X_{i,m}^{t}\oplus V_i))\setminus R_i^t
\end{eqnarray*}
We have $Z_{i}^t=(X_{i,m}^{t}\oplus V_i)\setminus R_i^t$
\begin{eqnarray*}
Z_i^{t+1/2}&=&Z_{i}^t+(A^t \oplus V_i)\setminus (X_{i,m}^{t}\oplus V_i)\setminus R_i^t
\end{eqnarray*}
As $(A\setminus B)\setminus C = (A\setminus (B \setminus C))\setminus C$ then with $Z_{i}^t=(X_{i,m}^{t}\oplus V_i)\setminus R_i^t$  
\begin{eqnarray*}
Z_i^{t+1/2}&=&Z_i^{t+1/2}+(A^t\oplus V_i)\setminus Z_i^{t}\setminus R_i
\end{eqnarray*}

\subsection{Degrowth of the myself region}
If there is only the degrowth of the myself region: $X_{i,m}^{t+1/2}=X_{i,m}^t+A^t$ then
\begin{eqnarray}
Z_i^{t+1/2}&=&Z_i^t - ((A^t\oplus V_i)\setminus(X_{i,m}^{t+1/2}\oplus V_i))\setminus (Z_i^t)^c
\end{eqnarray}

\underline{Lemme}:\\
\[A \oplus V = ((A \cup B) \oplus V) \setminus ( (B\oplus V)\setminus (A \oplus V)) \]
First, demonstrate  that $A = (A\cup B)\setminus (B\setminus A)$
 \begin{eqnarray*}
 A &=& A\cup \emptyset\\
A &=& A\cup (B\cap B^c)\\
A &=& (A\cup B)\cap (A\cup B^c)\\
A &=& (A\cup B)\cap (B^c\cup A )\\
A &=& (A\cup B)\setminus (B^c\cup A )^c\\
A &=& (A\cup B)\setminus (B\cap A^c )\\
A &=& (A\cup B)\setminus (B\setminus A)
\end{eqnarray*}
Switching $A$  by $ A\oplus V$ and $B$  by $ B\oplus V$ in this last formul, we get:
 \begin{eqnarray*}
A \oplus V = ((A\oplus V) \cup (B\oplus V)  )\setminus ( (B\oplus V)\setminus (A \oplus V))
 \end{eqnarray*}
As $ (A\oplus V ) \cup ( B\oplus V)=(A \cup B) \oplus V$, we find
 \begin{eqnarray*}
A \oplus V = ((A \cup B) \oplus V )\setminus ( (B\oplus V)\setminus (A \oplus V))
 \end{eqnarray*}

\textbf{Proof}:\\
\begin{eqnarray*}
Z_i^{t+1/2}&= & (X_{i,m}^{t+1/2} \oplus V_i)\setminus (\bigcup\limits_{j \in N_i} X^{t+1/2}_{j,o})
\end{eqnarray*}
As we have only the degrowth of the myself region, $X_{i,m}^{t+1/2}=X_{i,m}^t-A^t$ with $R_i^t=\bigcup\limits_{j \in N_i} X^{t}_{j,o}$ then
\begin{eqnarray*}
Z_i^{t+1/2}&= & ((X_{i,m}^t-A^t) \oplus V_i)\setminus R_i^t
\end{eqnarray*}
By the lemme, switching $A$ by $(X_{i,m}^{t}-A^t)$ and  $B$ by $A^t$, we have
\begin{eqnarray*}
Z_i^{t+1/2}&=& (A_1 \setminus A_2 )\setminus R_i^t\\
&\mbox{with}& A_1 = ((X_{i,m}^{t}-A^t)\cup A^t)\oplus V_i \\
&\mbox{with}& A_2 = (A^t\oplus V_i)\setminus((X_{i,m}^{t}-A^t)\oplus V_i)
\end{eqnarray*}
Like $((X_{i,m}^{t}-A^t)\cup A^t = X_{i,m}^{t}$ so we obtain $A_1=X_{i,m}^{t} \oplus V_i $.\\
As $X_{i,m}^{t}-A^t=X_{i,m}^{t+1/2}$ so $A_2 = (A^t\oplus V_i)\setminus(X_{i,m}^{t+1}\oplus V_i)$. We have
\begin{eqnarray*}
Z_i^{t+1/2}&=& (A_1 \setminus A_2 )\setminus R_i^t\\
&\mbox{with}& A_1 = X_{i,m}^{t}\oplus V_i \\
&\mbox{with}& A_2 = (A^t\oplus V_i)\setminus(X_{i,m}^{t+1/2}\oplus V_i)
\end{eqnarray*}
As $(A\setminus B)\setminus C = (A\setminus C)\setminus B$, substituting $A$ by $A_1$, $B$ by $A_2$ and $C$ by $R_i^t$ in this last formula. As $A_1\setminus R_i^t=Z_i^{t}$, we have
\begin{eqnarray*}
Z_i^{t+1/2}=&Z_i^{t} \setminus A_2\\
\mbox{with}& A_2 = (A^t\oplus V_i)\setminus(X_{i,m}^{t+1/2}\oplus V_i)
\end{eqnarray*}
As $A\setminus B= A - (B\setminus A^c)$, we have:
\begin{eqnarray*}
Z_i^{t+1/2}&=&Z_i^t - ((A^t\oplus V_i)\setminus(X_{i,m}^{t+1/2}\oplus V_i))\setminus (Z_i^t)^c
\end{eqnarray*}

\subsection{Growth of the other region}
If there is only the growth of the other region: $X_{i,o}^{t+1}=X_{i,o}^{t+1/2}+A^t$ then
\begin{eqnarray}
\forall j \in \mathbb{N}_n\begin{cases} Z_j^{t+1}=Z^{t+1/2}_j\\ - A^t\setminus (Z^{t+1/2}_j)^c &\mbox{ if } i \in N_j\wedge V_j\neq \emptyset \\ Z_j^{t+1}=Z^{t+1/2}_j &\mbox{ else}\end{cases}
\end{eqnarray}
\textbf{Proof:}\\
Let $V_j=\emptyset$, then  $\forall t : Z_j^{t}=\emptyset$, in particulary $Z_j^{t+1}=Z^{t+1/2}_j$.
Let assume now that $V_j=V$. 
\begin{eqnarray*}
Z_j^{t+1}=&(X^{t+1}_{j,m}\oplus V)\setminus \cup_{k\in N_j} X^{t+1}_{k,o}
\end{eqnarray*}
By the commutativity of union, we have:
\begin{eqnarray*}
Z_j^{t+1}=&(X^{t+1}_{j,m}\oplus V)\setminus( \cup_{k\in N_j\setminus i} X^{t+1}_{k,o} \cup
X^{t+1}_k)
 &\mbox{ if } i \in N_j\\
Z_j^{t+1}=&Z_j^{t} &\mbox{ else} 
\end{eqnarray*}
We suppose $i\in N_j$. We have only the growth of the other region: $X_{i,o}^{t+1}=X_{i,o}^{t+1/2}+A^t$ 
\begin{eqnarray*}
Z_j^{t+1}=&(X^{t+1/2}_{j,m}\oplus V)\setminus( \cup_{k\in N_j\setminus i} X^{t+1/2}_{k,o} \cup
(X^{t+1/2}_k+A^t))
\end{eqnarray*}
$A+B=A\cup B$ and by the commutativity of union, we have:
\begin{eqnarray*}
Z_j^{t+1}&=&(X^{t+1/2}_{j,m}\oplus V)\setminus (\cup_{k\in N_j} X^{t+1/2}_{k,o} \cup A^t)
\end{eqnarray*}
As  $A\setminus (B\cup C)=A\setminus B\setminus C$ and $Z_j^{t+1/2}=(X^{t+1/2}_{j,m}\oplus V)\setminus (\cup_{k\in N_j} X^{t+1/2}_{k,o})$, we have
\begin{eqnarray*}
Z_j^{t+1}&=&Z_j^{t+1/2}\setminus A^t
\end{eqnarray*}
As $A\setminus B= A - B\setminus A^c$, thus
\begin{eqnarray*}
Z_j^{t+1}&=&Z_j^{t+1/2}- A^t\setminus (Z_j^{t+1/2})^c
\end{eqnarray*}

\subsection{Degrowth of the other region}
If there is only the growth of the other region:$X_{i,o}^{t+1}=X_{i,o}^{t+1/2}-A^t$ then
\begin{eqnarray*}
\forall j \in \mathbb{N}_n\begin{cases} Z_j^{t+1}=Z_j^{t+1/2}+\\(((A^t\setminus (X^{t+1/2}_{m,j}\oplus V_j)^c)\setminus  (\bigcup\limits_{k \in N_j} X^{t+1}_{k,o}))\setminus Z_j^{t+1/2})\\ \hspace{2cm}\mbox{ if } (i \in N_j) \wedge (V_j\neq \emptyset) \\ Z_j^{t+1}=Z_j^{t+1/2}  \mbox{ else}\end{cases}
\end{eqnarray*}
\textbf{Proof:}\\
Let $V_j=\emptyset$, then  $\forall t :Z_j^{t}=\emptyset$, in particulary $Z_j^{t+1}=Z^{t+1/2}_j$.
Let assume now that $V_j=V$. 
Let $j$ in $\mathbb{N}_n$
\begin{eqnarray*}
Z_j^{t+1}=&(X^{t+1}_{j,m}\oplus V)\setminus \cup_{k\in N_j} X^{t+1}_{k,o}
\end{eqnarray*}
By the commutativity of union, we have:
\begin{eqnarray*}
Z_j^{t+1}=&(X^{t+1}_{j,m}\oplus V)\setminus( \cup_{k\in N_j\setminus i} X^{t+1}_{k,o} \cup
X^{t+1}_k)
 &\mbox{ si } i \in N_j\\
Z_j^{t+1}=&Z_j^{t+1/2} &\mbox{ if} i \notin N_j
\end{eqnarray*}
We suppose $i\in N_j$. We have only the growth of the other region: $X_{i,o}^{t+1}=X_{i,o}^{t+1/2}-A^t$
\begin{eqnarray*}
Z_j^{t+1}=&(X^{t+1/2}_{j,m}\oplus V)\setminus( \cup_{k\in N_j\setminus i} X^{t+1/2}_{k,o} \cup
(X^{t+1/2}_{i,o}-A^t))
\end{eqnarray*}%
As $A-B=A\cap B^c$, we have
\begin{eqnarray*}
Z_j^{t+1}=&(X^{t+1/2}_{j,m}\oplus V)\setminus( \cup_{k\in N_j\setminus i} X^{t+1/2}_{k,o} \cup
(X^{t+1/2}_{i,o}\cap(A^t)^c))
\end{eqnarray*}
As $A\cup (B\cap C)= (A\cup B)\cap(A\cup C)$, thus
\begin{eqnarray*}
Z_j^{t+1}=&(X^{t+1/2}_{j,m}\oplus V)\\
&\setminus( \cup_{k\in N_j} X^{t+1/2}_{k,o} \cap(\cup_{k\in N_j\setminus i} X^{t+1/2}_{k,o}\cup (A^t)^c))
\end{eqnarray*}
As $A \setminus (B \cap C)= (A\setminus B)\cup (A\setminus C)$, thus:
\begin{eqnarray*}
Z_j^{t+1}=&Z_j^{t+1/2}\cup A_1\\
\mbox{with } &A_1=(X^{t+1/2}_{j,m}\oplus V) \setminus( \cup_{k\in N_j\setminus i} X^{t+1/2}_{k,o}\cup (A^t)^c)
\end{eqnarray*}

As $A\setminus (B \cup C)=(B^c \cap A) \setminus C$, we have
\begin{eqnarray*}
A_1=(A^t\cap  (X^{t+1/2}_{j,m}\oplus V) ) \setminus(\cup_{k\in N_j\setminus i} X^{t+1/2}_{k,o} )\\
A_1=(A^t\setminus  (X^{t+1/2}_{j,m}\oplus V)^c ) \setminus(\cup_{k\in N_j\setminus i} X^{t+1/2}_{k,o} )\\
A_1=(A^t\setminus  \cup_{k\in N_j\setminus i} X^{t+1/2}_{k,o}  ) \setminus( (X^{t+1/2}_{j,m}\oplus V)^c)
\end{eqnarray*}
As $X_{i,o}^{t+1}=X_{i,o}^{t+1/2}-A^t$, so $A^t\cap X_{i,o}^{t+1/2}=\emptyset$ and $A^t=A^t\setminus X_{i,o}^{t+1} $. As $(A\setminus B)\setminus C=A\setminus (B\cup C)$, we have
\begin{eqnarray*}
A_1=&(A^t\setminus ( \cup_{k\in N_j\setminus i} X^{t+1/2}_{k,o}\cup X_{i,o}^{t+1} ) ) \setminus( (X^{t+1/2}_{j,m}\oplus V)^c)\\
A_1=&(A^t\setminus ( \cup_{k\in N_j} X^{t+1}_{k,o}  ) \setminus( (X^{t+1/2}_{j,m}\oplus V)^c)\\
A_1=&(A^t\setminus (X^{t+1/2}_{j,m}\oplus V)^c   \setminus( \cup_{k\in N_j} X^{t+1}_{k,o})
\end{eqnarray*}
The last step is: $Z_j^{t+1/2}\cup A_1= Z_j^{t+1/2}+ A_1\setminus Z_j^{t+1/2}$.

% \section{The hidden implementation}
% A compléter
% \label{implementation}
% The hidden implementation is the actualization and the link between the SQ. During the algorithm, the regions fluctuations imply fluctuation of the ZI. In the subsection~\ref{toto}, we give a procedure to actualize the ZI efficiently.  During this actualization, each time there is a growth, $x$, to the ZI $i$, we attribute the couple $(x,i)$ in the queue $\delta(x,i)$.  
% 
% \section{determinist  Watershed transformation }
% \label{dw}
% A faire 
% \bibliographystyle{plain}
% \bibliography {../bibliogenerale}
%  \end{document}

\section*{Acknowledgment}
I would like to thank my Ph.d supervisor, P. Levitz, for his support and his trust. The author is indebted to P. Calka for valuable discussion and C. Wiejak for critical reading of the manuscript.  I express my gratitude to the Association Technique de l'Industrie des Liants Hydrauliques (ATILH) and the French ANR project "mipomodim" No. ANR-05-BLAN-0017 for their financial support.

\bibliographystyle{plain}
\bibliography {bibliogenerale}

\begin{thebibliography}{1}

\bibitem{ADAMS1994}
R.~Adams and L.~Bisschof.
\newblock Seeded region growing.
\newblock {\em Ieee Transactions On Pattern Analysis And Machine Intelligence},
  16(6):641--647, June 1994.

\bibitem{Ballard1982}
D.H. Ballard and C.~Brown.
\newblock {\em Computer Vision}.
\newblock Berlin, Germany: Springer Verlag, 1982.

\bibitem{Beucher1979}
S.~Beucher and C.~Lantuejoul.
\newblock Use of watersheds in contour detection.
\newblock In {\em real-time edge and motion detection}. International workshop
  on image processing, 1979.

\bibitem{Hojjatoleslami1998}
S.~A. Hojjatoleslami and J.~Kittler.
\newblock Region growing: A new approach.
\newblock {\em Ieee Transactions On Image Processing}, 7(7):1079--1084, July
  1998.

\bibitem{KANADE1994}
T.~Kanade and M.~Okutomi.
\newblock A stereo matching algorithm with an adaptive window - theory and
  experiment.
\newblock {\em Ieee Transactions On Pattern Analysis And Machine Intelligence},
  16(9):920--932, September 1994.

\bibitem{Serra1982}
J.~Serra.
\newblock {\em Image Analysis and Mathematical Morphology - Vol. I . 610 p.}
\newblock Ac. Press, London, 1982.

\bibitem{Tariel2008d}
V.~Tariel.
\newblock Conceptualization of seeded region growing by pixels aggregation.
  part 3: a wide range of algorithms.
\newblock {\em submitted}, 2008.

\bibitem{Tariel2008f}
V.~Tariel.
\newblock Conceptualization of seeded region growing by pixels aggregation,
  part 6: random growing proceess.
\newblock {\em submitted}, 2008.

\end{thebibliography}
\end{document}